\documentclass[10pt,twocolumn,letterpaper]{article}

\usepackage[pagenumbers]{cvpr}            %

\usepackage{microtype}

\usepackage{placeins}
\usepackage{multirow}
\usepackage{dirtytalk}
\usepackage{cuted}
\usepackage{capt-of}

\definecolor{cvprblue}{rgb}{0.21,0.49,0.74}
\usepackage[pagebackref,breaklinks,colorlinks,allcolors=cvprblue]{hyperref}

\title{World Machine: Towards Generative World Modeling for Time-Series}

\author{
\textbf{Elton Cardoso do Nascimento}\textsuperscript{1} \qquad \textbf{Alexandre da Silva Simões}\textsuperscript{2} \qquad \textbf{Esther Luna Colombini}\textsuperscript{3} \\ \textbf{Ricardo Ribeiro Gudwin}\textsuperscript{1} \qquad \textbf{Paula Dornhofer Paro Costa}\textsuperscript{1} \\
\textsuperscript{1, 3}Universidade Estadual de Campinas (UNICAMP), \{FEEC, IC\}\\ \textsuperscript{2}Universidade Estadual Paulista (UNESP), Instituto de Ciência e Tecnologia, Sorocaba \\
{\tt\small e233840@dac.unicamp.br, alexandre.simoes@unesp.br, \{estherlc, gudwin, paulad\}@.unicamp.br}
\vspace{-5mm}
}

\begin{document}
\maketitle
\begin{strip}
    \centering
    \includegraphics[width=\linewidth]{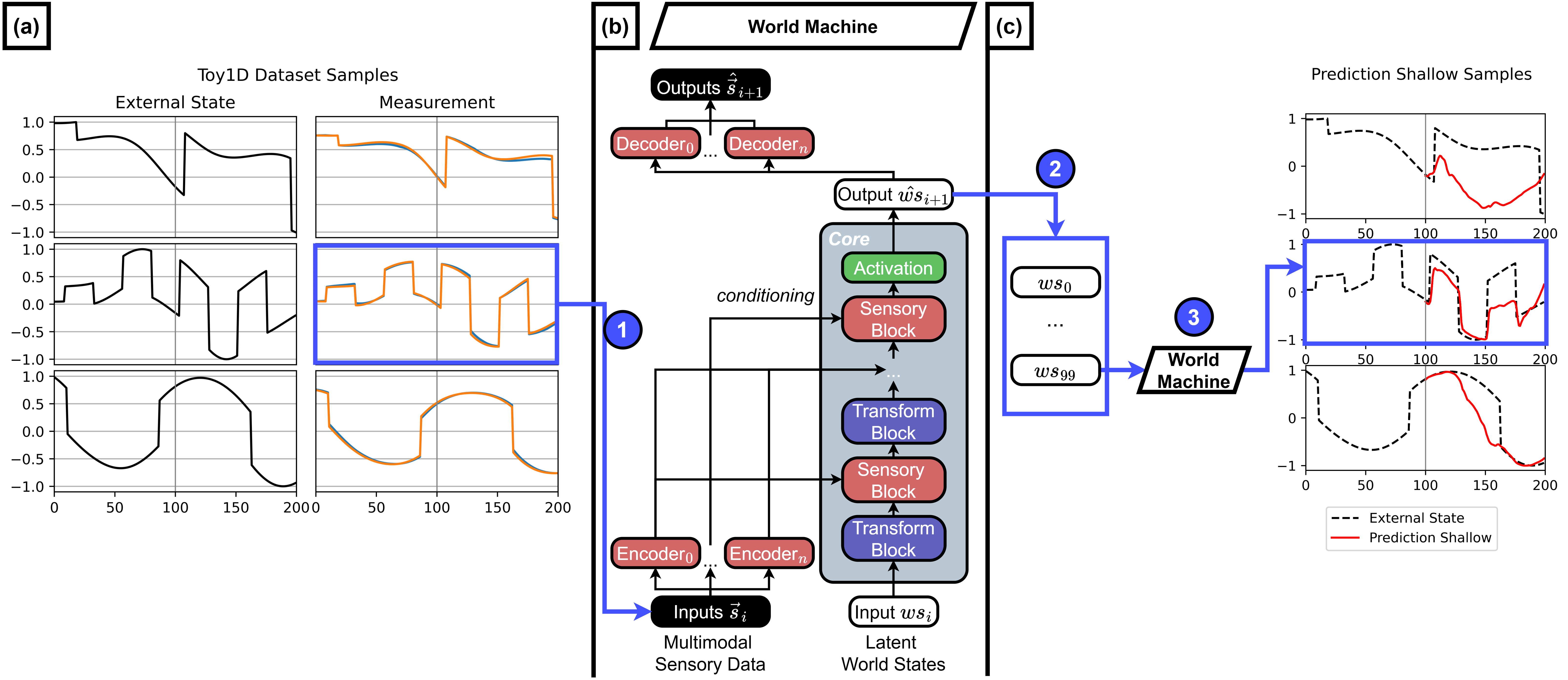}
    \captionof{figure}{\emph{World Machine operation in the Prediction Shallow Task}. The Toy1D dataset  \protect\say{measurement}  (a) is sent to the World Machine as an input sensory channel (1), increasing one piece of data at each temporal step, until reaching 50\% of the sequence in 100 steps. The model autoregressively performs inference up to this 50\% of the sequence (2), generating the sequence of latent world states $ws$ (c). After, using only the last state, $ws_{99}$, infers the last 50\% of the data, decoding the state to generate the output sensory channel \protect\say{external state} (c). We selected the samples from this figure randomly.
 }
    \label{fig:main_figure}
\end{strip}

\begin{abstract}

World models represent a paradigm shift in generative AI, pursuing predictive understanding and controllable simulation of environments in a structured and generalizable way. We present \textbf{World Machine}, a generative world-modeling architecture for time series. It is a transformer-based architecture with latent states that enables adaptation to different amounts of observed data and contexts. This shows an improvement over traditional transformers, which have a computational and memory cost that scales quadratically with the context. Experiments on a proposed synthetic dataset, Toy1D, validate the approach's feasibility, demonstrate capabilities not found in conventional transformers, and highlight the contributions of each component of the training protocol.
\end{abstract}

\section{Introduction}
\label{sec:intro}

Deep learning-based world models are an emergent new category of models. They exhibit behaviors that can be used to simulate the environments in which agents can learn and act, allowing us to explore and play in them, or to create agents that better understand and act autonomously in our world \cite{Hafner2020Dream, genie3, lecunPathAutonomousMachine}. 

To intuitively understand what a world model is, suppose a volleyball player. He observes the ball rising beside the net, and his brain rapidly \textbf{integrates multiple streams of sensory information}—including vision, hearing, and proprioception. This perceptual data is immediately combined with prior experience: his knowledge of the rules of the game, the \say{embodied physical intuition} developed through countless hours of practice, and the \textbf{affective memories} associated with these experiences. Through this complex synthesis, he \textbf{constructs an understanding} of the entities in his environment—the other players, the net, and the ball—and \textbf{anticipates} their trajectories in the moments ahead. Based on these predictions, he \textbf{formulates and selects} a course of action. As he executes this plan, he continuously \textbf{updates his perception in real time}, adjusting his movements dynamically until the play is completed.
This entire process occurs in a few seconds, making it impossible to respond in time to environmental stimuli without the brain's ability to predict future states of the world, based solely on present sensory inputs. This internal imaginary representation of the world is a \textbf{world model}. 

Computational world models may represent a promising solution for applications that require more \textbf{adaptive} systems—models capable of contextually integrating within complex environments. They can also enable the development of \textbf{generalist} models, encompassing a broad range of configurations with enhanced realism.  They can assist in \textbf{planning} tasks, where they need to predict the sequence of actions an agent can perform, as well as \textbf{reactively} respond to changes. They can also be capable of handling \textbf{heterogeneous data}, with different formats, dynamics, and frequencies.

In this work, we propose and examine the fundamental components of the \textit{World Machine}, a novel architecture and training protocol for computational world models, designed to explore the potential of world model-based deep learning approaches.
 Among the characteristics of this architecture are the creation of internal representations of the world model, the ability to handle different sensory channels (potentially across multiple modalities), and a controllable trade-off between quality and cost of predictions. We present specific differences between our approach and other current approaches in \cref{sec:related_work}.

We evaluate the proposed approach on a simple synthetic dataset using tasks designed to assess the model's specific capabilities. We demonstrate how the approach enables inference with context truncation, albeit at reduced performance, thereby decreasing the quadratic cost associated with current transformers. 

The code for the proposed architecture, protocol, and experiments are released \cite{cardosodonascimentoWorldMachine2025}, as well as the Docker image containing the environment used in this work \cite{cardosodonascimentoWorldMachineDocker2025}.

\section{Related Work}
\label{sec:related_work}

Learning the dynamics of three-dimensional digital game worlds is the focus of several recent models, such as GameNGen, Oasis, and WHAM~\cite{valevski2025diffusion, decartOasisUniverseTransformer2024, kanervistoWorldHumanAction2025}. These models are trained on gameplay data—comprising images and player actions—to develop generative systems that capture and reproduce the underlying dynamics of the games. In other words, by simply observing successive scenes of a game, as well as the actions performed by players at each instant, these models are trained to infer the rules of the game, the function of specific entities presented in the scenes, and, ultimately, the physics of the player's interactions in this virtual world. By doing so, they demonstrate the potential to perform capabilities commonly associated with hand-crafted simulations, as traditional game engines based on programmed rules. %

On a much larger scale, RFTM (Real-Time Frame Model) and Genie 3 are models capable of representing larger and more dynamic worlds, providing users with greater control \cite{worldlabsteamRTFMRealTimeFrame, genie3}. However, as the technical details of these systems are not yet fully available, it is difficult to analyze their underlying mechanisms or build upon their architectures. The limited information released indicates that these models also convert observable data into internal world representations and vice versa.

In terms of architecture, Dreamer \cite{Hafner2020Dream} creates a latent state to model the world. It performs this process using two models: one that learns the representation of the current state based on the previous state, action, and observation, and another that learns the transition from a previous state based on an action.

In this work, we propose a novel architecture that synthesizes and improves upon prior approaches. Similar to recent generative systems such as GameNGen, Oasis, and WHAM, our architecture models the world as an autoregressive sequence. However, our primary contribution is to explicitly model the internal state that generates this data, rather than correlations among sensory streams alone. In contrast to Dreamer, we adopt a single end-to-end architecture that unifies representation and transition, treats experience as a higher-order Markov process, and does not conceptually distinguish between channels such as actions and observations. The entire model is optimized with a single sensory loss. Finally, to address the opacity of recent large-scale models, we release a fully open-source, reproducible baseline.

\section{Method}
\label{sec:method}

We introduce the World Machine architecture and the proposed training protocol (\cref{sec:method-wm_and_protocol}). To evaluate the proposed approach and problem setting, we define a set of tasks and metrics (\cref{sec:method-evaluation}), as well as a synthetic dataset, \textit{Toy1D} (\cref{sec:method-dataset}). 

\subsection{World Machine and Train Protocol}
\label{sec:method-wm_and_protocol}

The \textbf{World Machine} is a generative, world modeling, multimodal model for time series (\cref{fig:main_figure} (b)). Generative as it generates data; world modeling as it tries to create an internal representation of the external world;  multimodal as it can operate simultaneously with multiple sensory types; and for time series as it generates and observes data in a sequential order. 

It is based on decoder-only transformer models that operate on \textbf{latent world states}.
A latent world state, denoted by $ws$, is a real-valued vector that encodes the current configuration of the simulated world. It plays a role analogous to the embedding representation in conventional transformer architectures. %
These latent states evolve over time and are organized sequentially at a fixed temporal resolution, referred to as the \textit{world simulation frequency}. 
The resulting sequence of world states forms a vector $\vec{ws}$, where each element corresponds to a specific temporal step as in
\begin{equation}
    \vec{ws} = [\, ws_0,\, ws_1,\, ws_2,\, \ldots,\, ws_{n-1} \,].
\end{equation}

\textbf{Sensory data} are the observations that the model makes of the external world. They are organized into sensory channels %
which can be input and/or output channels %
Each input channel %
passes through a pre-encoder, which generates a vector for each dimension at the temporal step. This comes with \textbf{sensory masks} indicating the presence or absence of the data in a temporal step. The $ws$ can be decoded to generate the sensory data for each channel %
in the current step.

The main part of the model, which actually performs the simulation, is called the \textbf{Core}. It is capable of generating the next state using the previous state, and optionally multiple previous states and sensory data. It consists of transformer blocks, which can be standard blocks without sensory conditioning, or sensory blocks that are conditioned by sensory data, each block receiving data from a specific channel.
The sensory blocks used were of the adaLn-Zero\cite{peeblesScalableDiffusionModels2023} type, which perform scales and shifts on the main data (embedding/latent world state) modulated by the conditioning (sensory) data. In the case of non-existent sensory data, the block does not perform these operations in the temporal step.

After all the blocks, the Core performs a \textbf{state activation} using $tanh$. As we discuss in \cref{sec:experiments-experiment1}, this operation is important for maintaining the model's stability. 

We use \emph{ALiBi} as positional encoding for two reasons, which set it apart from the commonly used sinusoidal encoder~\cite{press2022train}. First, because it is an unbound method, we can use sequences of different sizes. Second, because it uses a \emph{modifying Attention Matrix} injection method, the positional information is injected directly in the attention matrix, instead of in the embeddings/\emph{ws}, not adding unrelated disturbance to the $ws$ \cite{dufterPositionInformationTransformers2022}.

The model training task is to correctly predict the output sensory data for the next temporal step, analogous to a next token prediction task. To achieve this, the defined optimization loss is the weighted sum of the loss for each output sensory dimension. This creates a relationship between the $ws$ and the sensory data, as the $ws$ must be able to contain the necessary information to generate that data.

Training a World Machine requires a different process, since the Core operates on $ws$, this data is needed beforehand. However, the model itself figures out how to encode these states. It is comparable to trying to train a model without having all the input data. The \textbf{State Discovery} process was developed to solve this problem. The states are initialized at the beginning of training using a uniform distribution ($\sim Uniform(-1, 1)$). In each epoch, the predicted $\hat{\vec{ws}}$ in the Core output are temporally shifted, using the \say{shift to left operator} $\mathcal{S}$, and saved back as input data for the next epoch. We indicate the epoch of the latent state vector with an upper index, $\vec{ws}^j$. The first state in the sequence, in all epoch, is modified to be a null vector:
\begin{align}
    \vec{ws}^j =  \mathcal{S}(\hat{\vec{ws}}^{j-1})
    &&
    	\forall j, ws^j_0 = \vec{0}.
    \label{eq:method-wm_and_protocol-shift}
\end{align}
With that, while the training optimizes the model's weights, the $ws$ corresponding to each input in the dataset are also updated. During inference, the model operates similarly to a regular transformer, performing an autoregressive process that starts from a null vector.

\begin{figure}[hbt]
    \centering
    \includegraphics[width=0.75\linewidth]{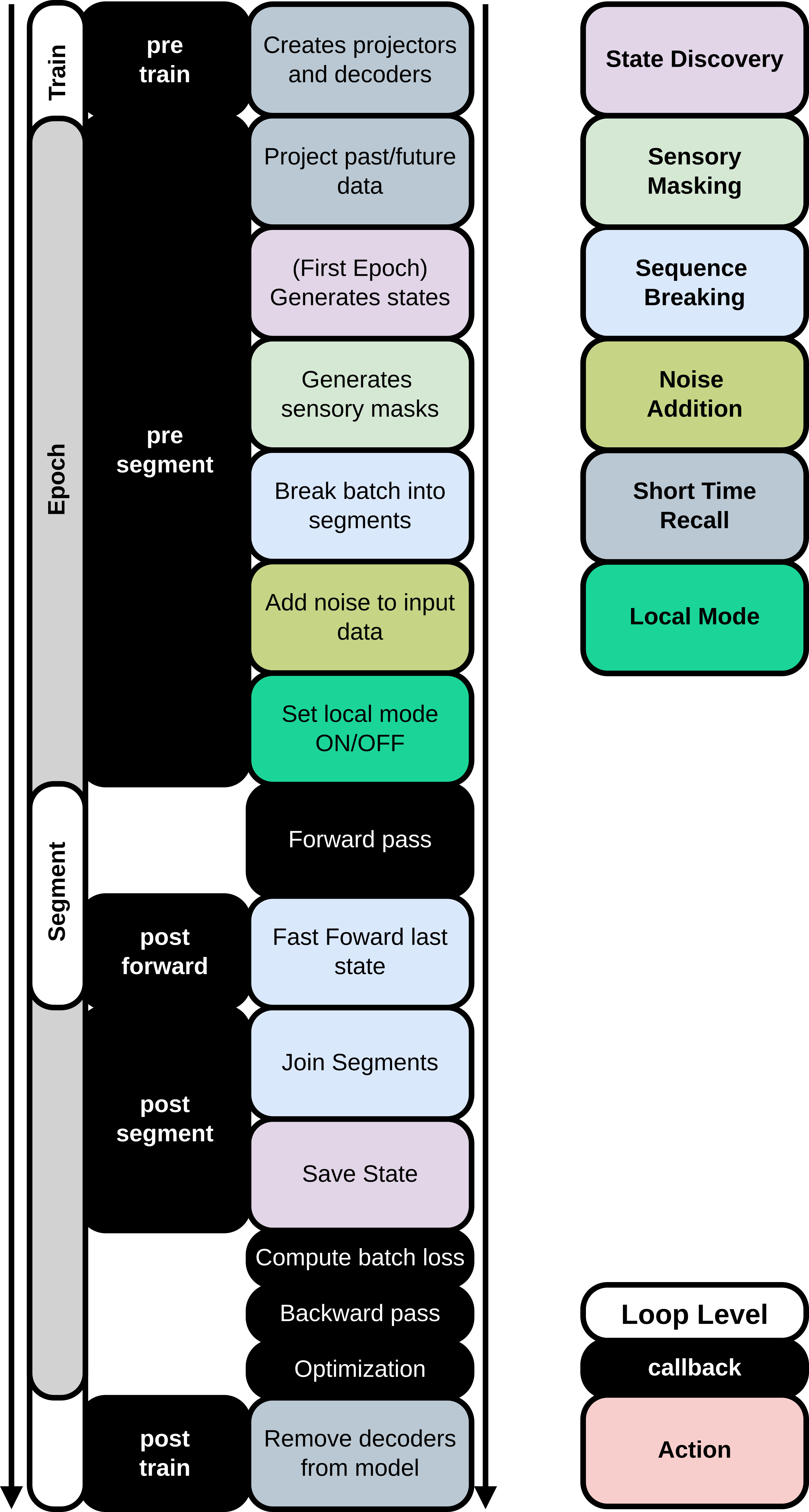}
    \caption{World Machine train process, when all protocols are used. It is structured in loops, with the addition of a ``segment loop'', created to iterate through the segments. Each loop has code callback points. The actions are organized by the protocols that implement them.}
    \label{fig:method-wm_and_protocol_trainloop}
\end{figure}

In order to improve the model and strengthen its ability to generate useful representations in the $ws$, we also propose other steps in the training process. \cref{fig:method-wm_and_protocol_trainloop} details the actions performed at each stage of the protocol described below:

\begin{itemize}
    \item \textbf{Sensory Masking:} involves masking sensory data during training, hiding a random amount of data at each batch. When masking, we can also verify the sensory masks before replacing the stored $ws$ with the new $ws$, ensuring that data is not replaced when it has been masked. This is particularly important at the beginning of training, when the model has not yet generated good states without sensory data. 
    
    \item \textbf{Sequence Breaking:} with this protocol, we break each sequence in random places, creating $k$ segments. Each segment passes through the model individually, and the outputs are concatenated to calculate the loss. Therefore, the model needs to be able to handle mid-sequence states without the preceding data. We can also shift the last state of one segment to the first state of the next, which we call \textbf{Fast Forward}. The hypothesis is that this increases the propagation of information from one state to another, helping the model learn to encode a good state, as the gradient propagates from the segment losses through the ``bottleneck'' generated by this single state. With an increase in the $k$ parameter, the training approaches a sequential autoregressive process.

    \item \textbf{Noise Addition:}  It adds Gaussian noise to the $ws$ and/or sensory dimensions in each batch, with controllable mean and standard deviation.
        
    \item \textbf{Short Time Recall:} This protocol creates new synthetic sensory channels that correspond to past and/or future data relative to the current temporal step. For each new channel, a new decoder must also be added to the model. Up to $n$ channels can be created, which are sampled using a stride. Each sampled data is projected by a random projector fixed at the start of the training.
    
    \item \textbf{Local Mode:} Randomly, for each segment, it can disable the model's attention process. The layer output is directly the value matrix after the external projection. This limits the model's vision, which needs to be able to handle only the current $ws$ and sensory data.
\end{itemize}

Both Sequence Breaking and Local Mode make the training step similar to training an RNN by processing the hidden state in parallel.

\subsection{Tasks and Metrics}
\label{sec:method-evaluation}

To evaluate the architecture and protocols created, we propose some distinct tasks to be performed after and that utilize the same type of loss as the training:

\begin{itemize}
    \item \textbf{Normal:} is a normal autoregressive inference. Note that this differs from the loss calculated during training, as we do not use states updated ``in parallel'' during the training process, but rather states estimated sequentially, starting from the null state.

    \item \textbf{Use State:} inference on previously encoded states, without sensory data. We can calculate this in one inference step, processing the elements of the sequence in parallel since the states have already been encoded. We only evaluate this task at the first 50\% of a sequence. If the model were merely taking sensory data and manipulating it to generate output, its performance on this task would be poor.

    \item \textbf{Prediction:} inference of future states, using several previous encoded states and without sensory data. We use the first 50\% of states to evaluate the task in the final 50\% of the sequence. 

    \item \textbf{Prediction Shallow:} inference of future states, using only one previous encoded state and without sensory data. We evaluate this task in the final 50\% of the sequence. This is the most important task, as it directly assesses the model's ability to perform inference with context truncation, which would otherwise incur a quadratic cost in sequence length and is a major issue with current transformers.
    
    \item \textbf{Prediction Local:} inference using local mode, that is, of next immediate state, using only one previous encoded state, without sensory data.

    \item \textbf{Mask Sensory@x}: inference in the full sequence randomly masking x\% of the sensory data.
\end{itemize}

Note that these tasks are not meaningful for a standard transformer model, as it is not capable of encoding latent representations suitable for continuing a temporal sequence. In such models, the individual token --- or, in our case, the sensory observation --- does not contain sufficient information to reliably predict the continuation of the sequence.

We also use an impact test to evaluate the impact of training hyperparameters on a model. Given a metric $M$ and a space of models trained with different binary hyperparameters, with a set A indicating the use of a specific hyperparameter, we define the impact as:
\begin{equation}
  I(M|A) := E[M|A] - E[M|\overline{A}].
  \label{eq:method-evaluation-impact}
\end{equation}
It is similar to some global model-agnostic interpretability methods, such as \emph{Leave One Feature Out} (LOFO), but differs in that it attempts to explain the model output in relation to input features \cite{molnar2025}. Here, we try to explain the model's overall performance in relation to the train hyperparameters 

We can also compute the impact relevance using a Wilcoxon signed-rank test. For that, we need models trained in both $A$ and $\overline{A}$ for each combination of other hyperparameters to generate \say{after adding hyperparameter A} and \say{before adding hyperparameter A} pairs.

\subsection{Toy1D Dataset}
\label{sec:method-dataset}

We proposed a synthetic dataset to understand how the World Machine behaves. We make this choice because a real dataset would generate unnecessary complexities, making it more difficult to understand the model's capabilities at this initial stage. The \emph{Toy1D} is a synthetic dataset of one-dimensional time series. The series represents a damped physical system, given by:
\begin{align}
    \vec{x}_{i+1} = F\vec{x}_i+\vec{u}_i
    &&
    F =  \begin{bmatrix} 
                1 & \Delta t & \frac{\Delta t^2}{2} \\
                -0.1 \Delta t & 1 & \Delta t \\
                0 & 0 & 1
            \end{bmatrix},
  \label{eq:method-dataset-system}
\end{align}
\noindent with random initial states $\vec{x}_0$. The $\Delta t$ is unitary. The initial $x_0$ of each series is random, and $\vec{u}_i$ is a random sum of square and impulse waves. The data's second and third elements, $x_i^1$, $x_i^2$, are clipped during the generation between -1 and 1 to avoid excessively high values. Only the position ($\vec{x}_i^0$) data is used in the final sensory dimension named \emph{external state}, with size 1. 

We define another sensory dimension, \emph{measurement}, with size 2, as:
\begin{align}
    \vec{s}_i = \tanh(H \vec{x}_i)
    &&
    H \sim Uniform(-1, 1)^{2\times2},
    \label{eq:method-dataset-measurement}
\end{align}

\noindent where $H$ is fixed in the start of the dataset generation. Note that, depending on the data scales and the H matrix, the measurement can become very similar to the external state, as can be seen in \cref{fig:main_figure} (a).

Since the dataset is stochastic, different data can be generated by controlling the seed of the random number generator. For each seed, we first generate 10,000 sequences of length 1,000 and then segment them into 40,000 sequences of length 200. Each sequence is scaled to the interval $[-1,1]$. \cref{fig:main_figure} (a) shows representative samples from the dataset. The dataset is split into 60\% for training, 20\% for validation, and 20\% for testing. The dataset is available online \cite{cardosodonascimentoWorldMachineToy1D2025}.

\section{Experiments and Results}
\label{sec:experiments}

This section presents the experiments conducted to study the behavior of the World Machine, along with the results and observations obtained. First, we tested the feasibility of training the proposed architecture (\cref{sec:experiments-experiment0}). Then, we analyzed the impact of each step of the proposed training protocol (\cref{sec:experiments-experiment1}).

In all experiments, models are trained on the Toy1D dataset. Note that Toy1D is dependent on the seed used for its generation. In experiments with more than one model trained per variation, the seed is varied, and, by doing so, the dataset is also varied. 

We used the sum of the MSE of the external state and measurement as the optimization loss. Training is performed using AdamW and a cosine-annealing scheduler with warmup for 100 epochs. The models have two blocks, both of which are sensory and utilize measurement as a sensory dimension (unless stated otherwise). Each attention layer has four heads. The state has a size dimension of 128. We use a batch size of 256. The experiments were performed by training multiple models simultaneously using 3x Nvidia A100 80 Gb, but we emphasize that it can be performed on simpler hardware.

Additional results and analysis are available in our experiment reports \cite{cardosodonascimentoWorldMachineData2025, cardosodonascimentoWorldMachineData2025a, cardosodonascimentoWorldMachineData2025b}.

\subsection{Protocol Experiment}
\label{sec:experiments-experiment0}

The goal of the protocol experiment was to conduct an initial check of whether the model is trainable, utilizing the concept of state discovery, and to assess the potential impact of the training protocol on the various tasks defined. For that, we train different model configurations. Each configuration may vary in the number of protocol steps used: \textbf{Base}, using only state discovery; \textbf{Sensory Mask}, using state discovery and sensory masking; \textbf{Complete Protocol}, using all steps, state discovery, sensory masking, sequence breaker, state-check sensory, fast forward, short time recall, noise adding and local mode. For each configuration, we trained 15 models and evaluated them on the defined tasks using both MSE and SDTW (Soft Dynamic Time Warping)\cite{cuturiSoftDTWDifferentiableLoss2017} to enable a comprehensive comparison.

\begin{figure}[hbt]
  \centering
   \includegraphics[width=0.8\linewidth]{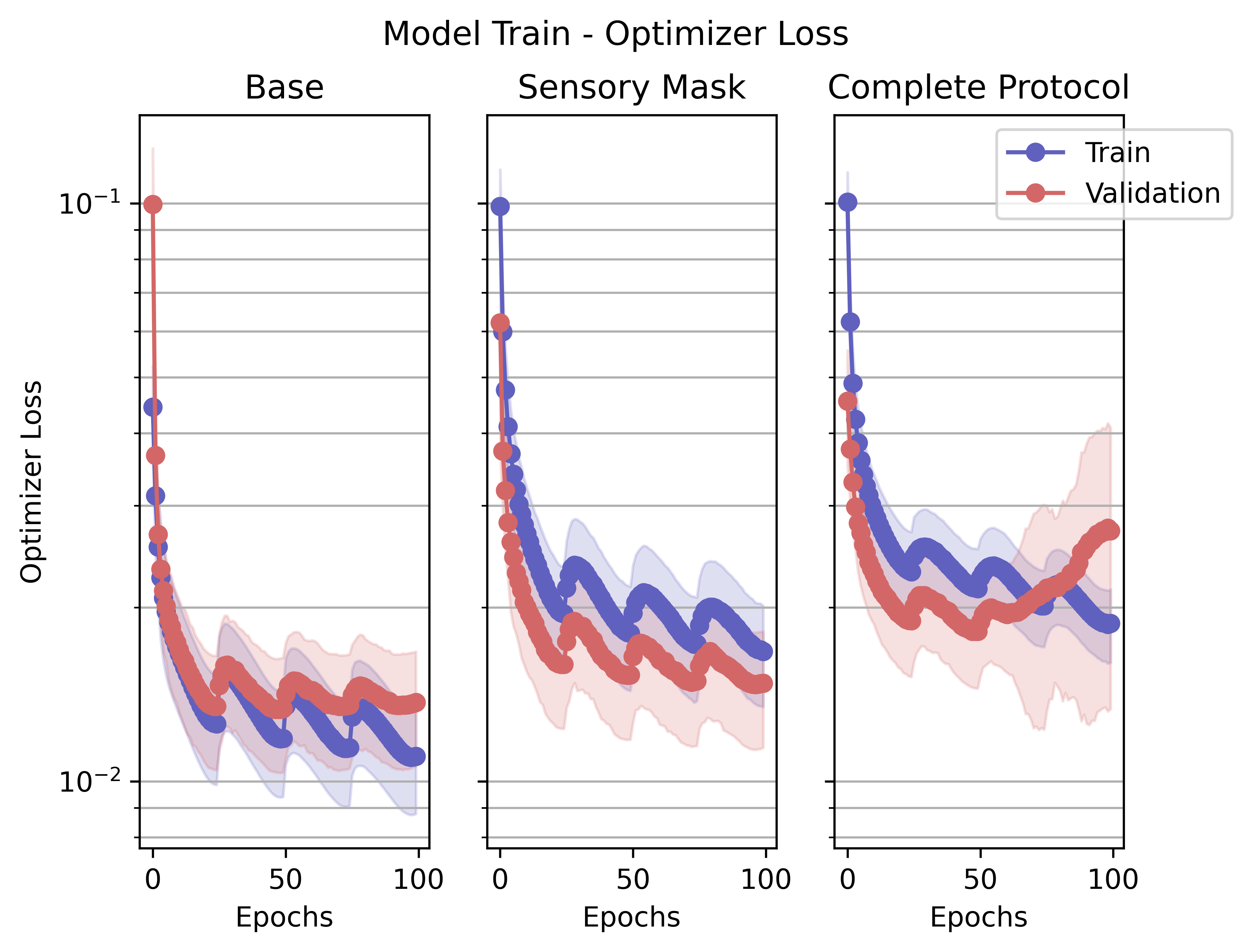}
   \caption{Loss of the optimizer of the models during the training of the protocol experiment.}
   \label{fig:experiments-experiment0-train_history}
\end{figure}

\Cref{fig:experiments-experiment0-train_history} shows the train and validation loss during training. We can see that the loss decreases when using the state discovery technique, indicating that the model is being successfully trained. We can also observe that, since we employ some of the protocol techniques in training but not in validation, this has an effect that tends to lead to a false conclusion of underfitting. Additionally, some training protocol steps are causing instability later in the training process.

\begin{figure}[hbt]
  \centering
    \begin{subfigure}{0.8\linewidth}
        \includegraphics[width=\linewidth]{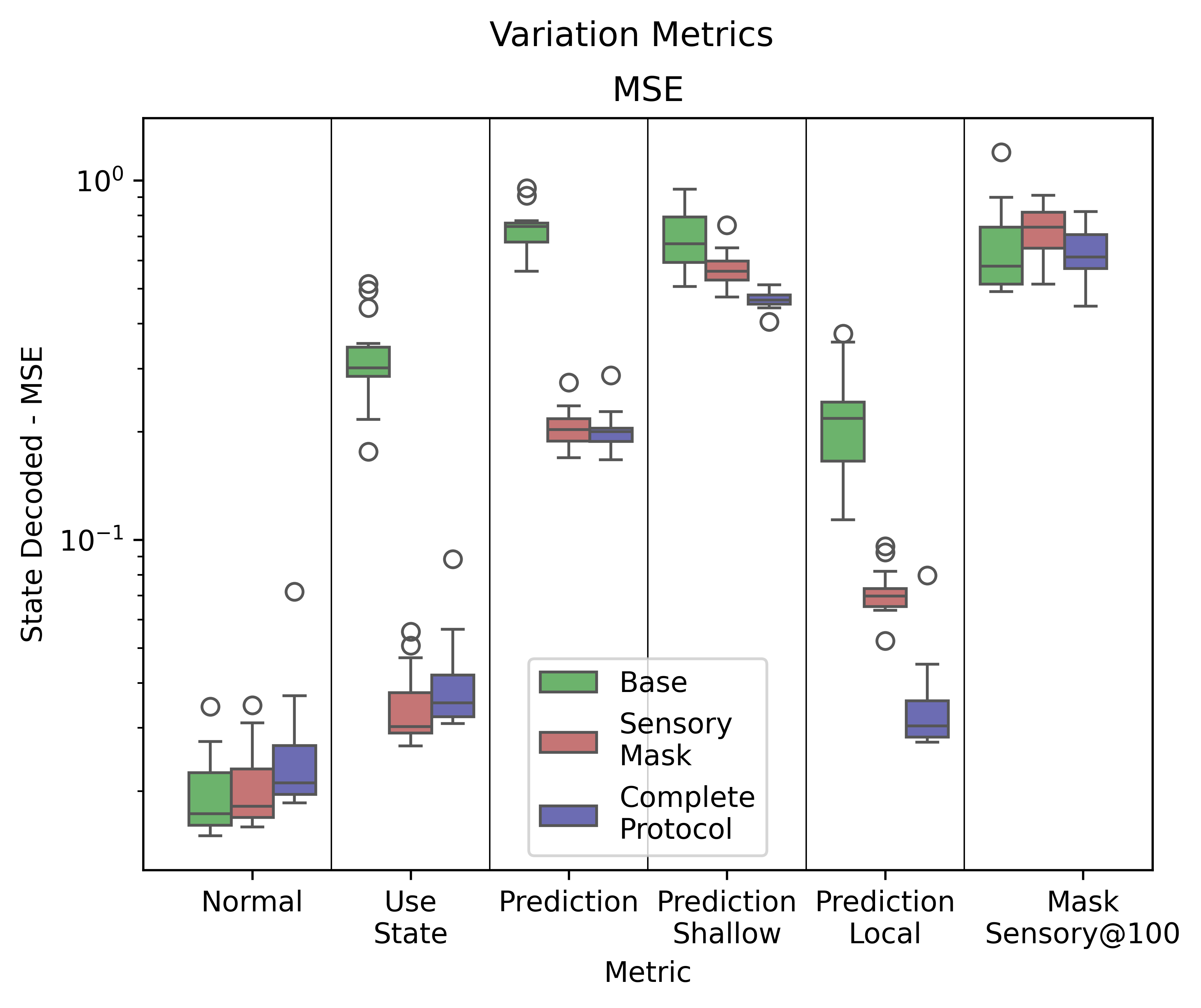}
        \caption{Task Performance using MSE criterion}
        \label{fig:experiments-experiment0-metrics-mse}
    \end{subfigure}

    \begin{subfigure}{0.8\linewidth}
        \includegraphics[width=\linewidth]{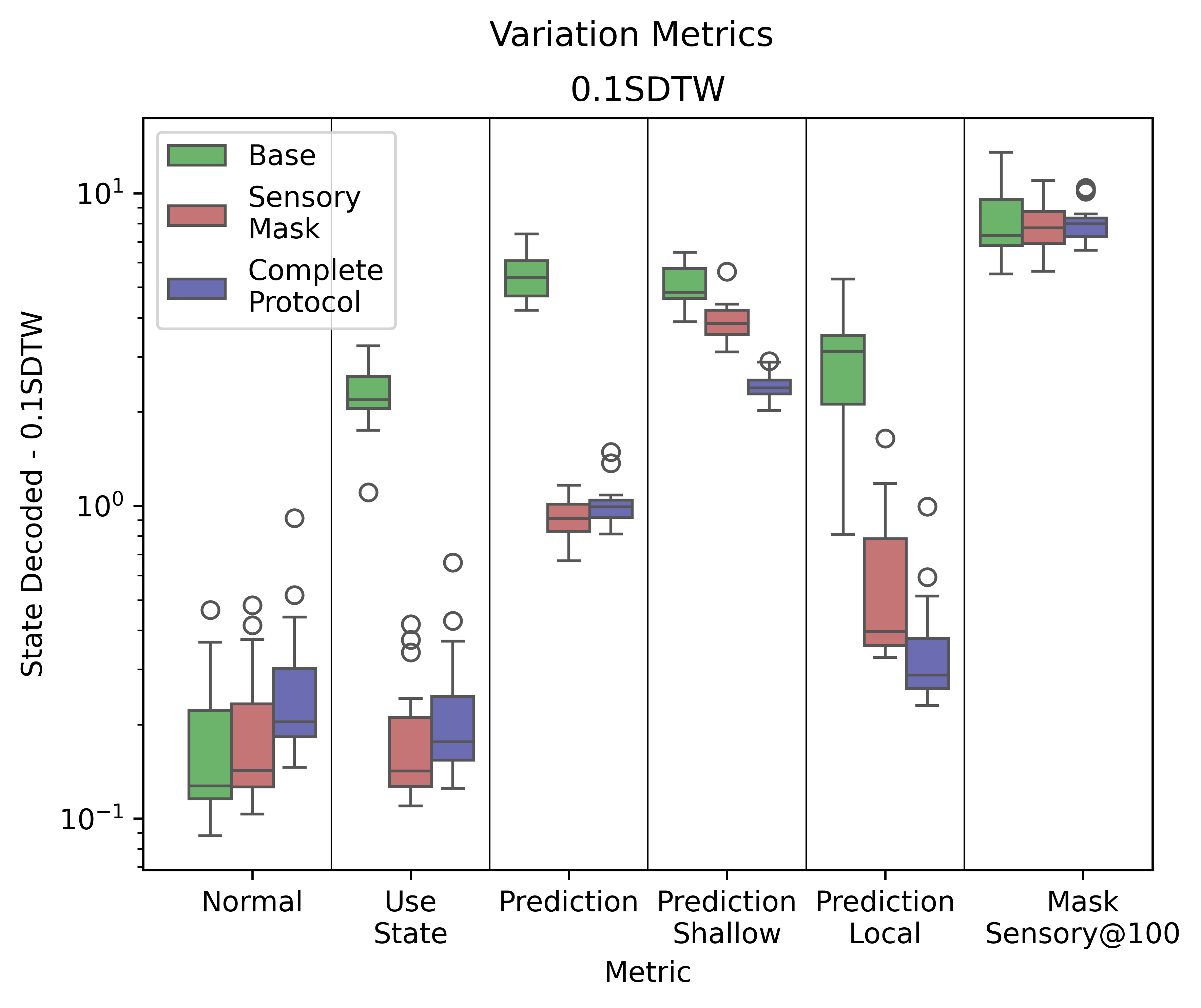}
        \caption{Task Performance using SDTW criterion}
        \label{fig:experiments-experiment0-metrics-sdtw}
    \end{subfigure}

    \caption{Model variations metrics in tasks}
    \label{fig:experiments-experiment0-metrics}
\end{figure}

Focusing our attention to the model performance in each task (\cref{fig:experiments-experiment0-metrics}), we can see that all models perform better in the normal task, but the Base model cannot perform other tasks that involve masking sensory data adequately, since it was not trained in the absence of sensory data. No model can simply make predictions in the absence of sensory data, since the task metrics are always lower than those in Mask Sensory@100. Mask Sensory@100 means attempting to generate a sequence without access to any sensory data, that is, without any information about what the sequence is like. Prediction Shallow is the most complex task to perform, as no model can perform it well, but the full protocol shows considerable improvements in this task. Comparing MSE with SDTW, SDTW captures more differences between variations, but it also generates similar conclusions to MSE, making it less important to analyze both metrics.

\begin{figure*}[hbt]
  \centering
   \includegraphics[width=0.8\linewidth]{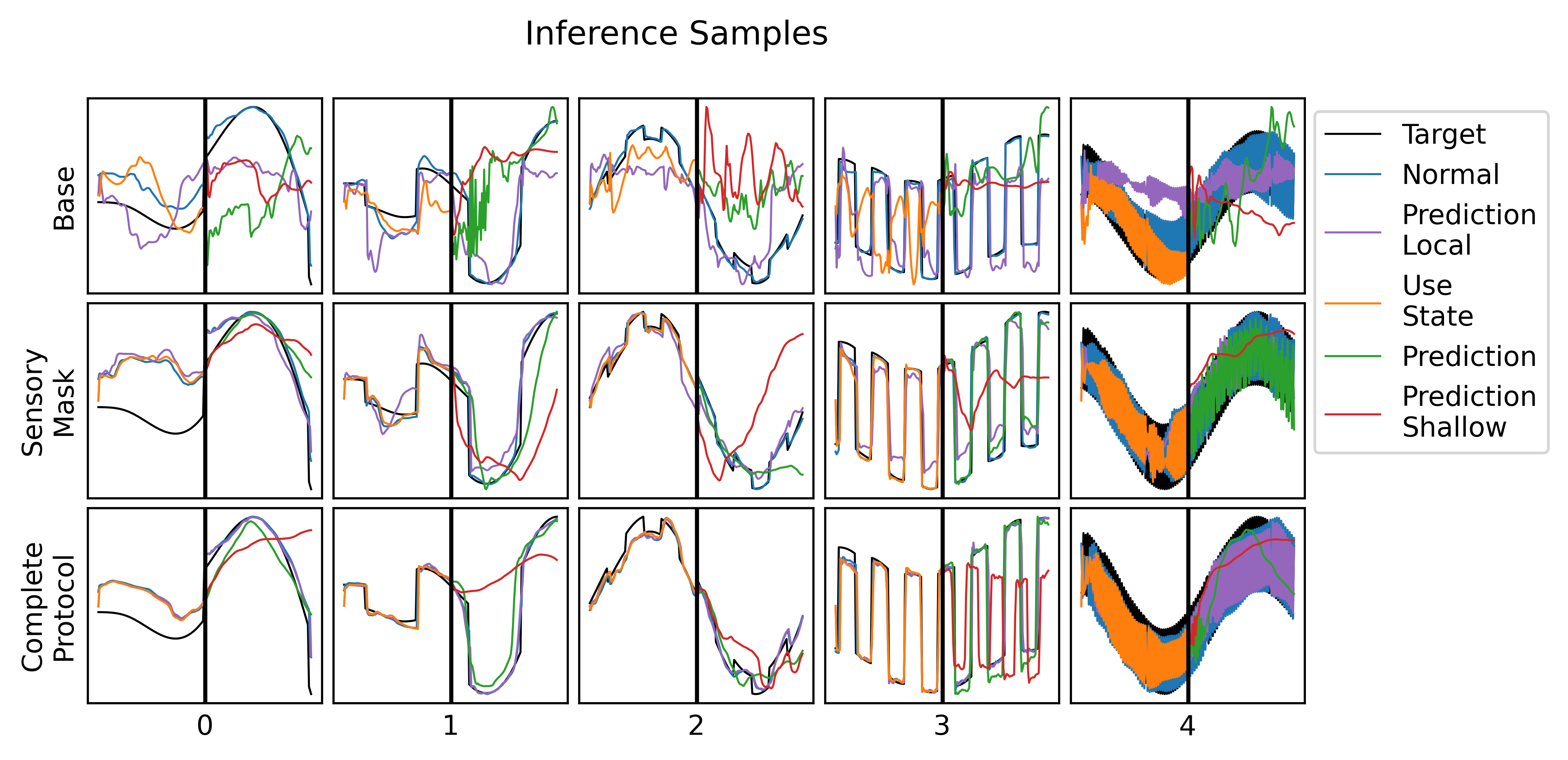}

   \caption{Model variations external state inference samples for four sequences in the validation dataset. The predictions of only one model for each variation are shown. The middle vertical bar indicates the middle of the sequence.}
   \label{fig:experiments-experiment0-samples}
\end{figure*}

In a qualitative view (\cref{fig:experiments-experiment0-samples}), we can see that the sequences that the model predicts are generally coherent, mainly without masking sensory data. Again, we can see that the Base model fails to predict coherent sequences in the absence of sensory data. Complete protocol can better follow the signal format (e.g., sequences 2 and 3), but still fails to predict future sequences correctly. High frequency sequences (e.g., sequence 4) may be more difficult to predict.

From these observations, we conclude that the proposed architecture is trainable using the proposed protocol, being capable of generating coherent predictions. The protocol is relevant to model performance, which varies across different tasks, with the Prediction Shallow task being the most challenging.

\subsection{Configuration Experiment}
\label{sec:experiments-experiment1}

Building on the previous experiment, which demonstrated the importance of the protocol for the model's final performance, we proceeded to investigate the specific impact of each step. For this purpose, we defined 15 indicator training variables that correspond to various combinations of hyperparameters. Some of these variables are mutually exclusive, meaning they never occur together. $\boldsymbol{SB_1}$ and $\boldsymbol{SB_2}$ are disjoint and corresponds to using sequence breaking, with $SB_2$ also using fast forwarding. $\boldsymbol{SM_1}$ corresponds to using a mean between the previous and new state estimates to define the next state, instead of only replacing the next state with the new state. $\boldsymbol{SM_2}$ is when we check the sensory data masks before updating the state. $\boldsymbol{AC_1}$ replaces the $\tanh$ state activation function for a MSE regularizer of the $ws$, added to the optimizer loss. $\boldsymbol{MD_1}$ is about changing the block configuration, from two blocks with the measurement as a sensory dimension to one block with the measurement and the second one as the $ws$ at the start of the step as a sensory dimension. $\boldsymbol{NA_1}$ is a noise addition in the state, and $\boldsymbol{NA_2}$ in the measurement dimension. $\boldsymbol{RF_{1-4}}$ and $\boldsymbol{RP_{1-4}}$ are variations of, respectively, future and past short-time recall, with varying stride and number of dimensions. Each variable in a group $RF$/$RP$ is disjoint from the others. $\boldsymbol{LM_1}$ corresponds to using local mode with $25\%$ chance per segment.

We trained only once for each variation corresponding to a combination of variables. All variations utilize sensory masking. Considering variables that are disjoint, we trained 9600 variations.

The first analysis we performed was the probability of a model diverging. We considered a model to diverge when, in any task, its metric is NaN or greater than three times the standard deviation of the metric, computed across all variation metrics without NaN values. We first observed a considerable chance of divergence $P(Diverge|Variable)$, near 20\% for almost every variable. But, when disregarding variations that used the variable $AC_1$, $P(Diverge|Variable \cap \overline{AC_1})$, the divergence probability turns into less than 1\% for every other variable. This showcases the importance of activating the state using $\tanh$ for better model stability.

For the following analyses, we will only consider the non-divergent models, which account for 84.01\% of the variations. 

\begin{figure}
    \centering
    \includegraphics[width=0.75\linewidth]{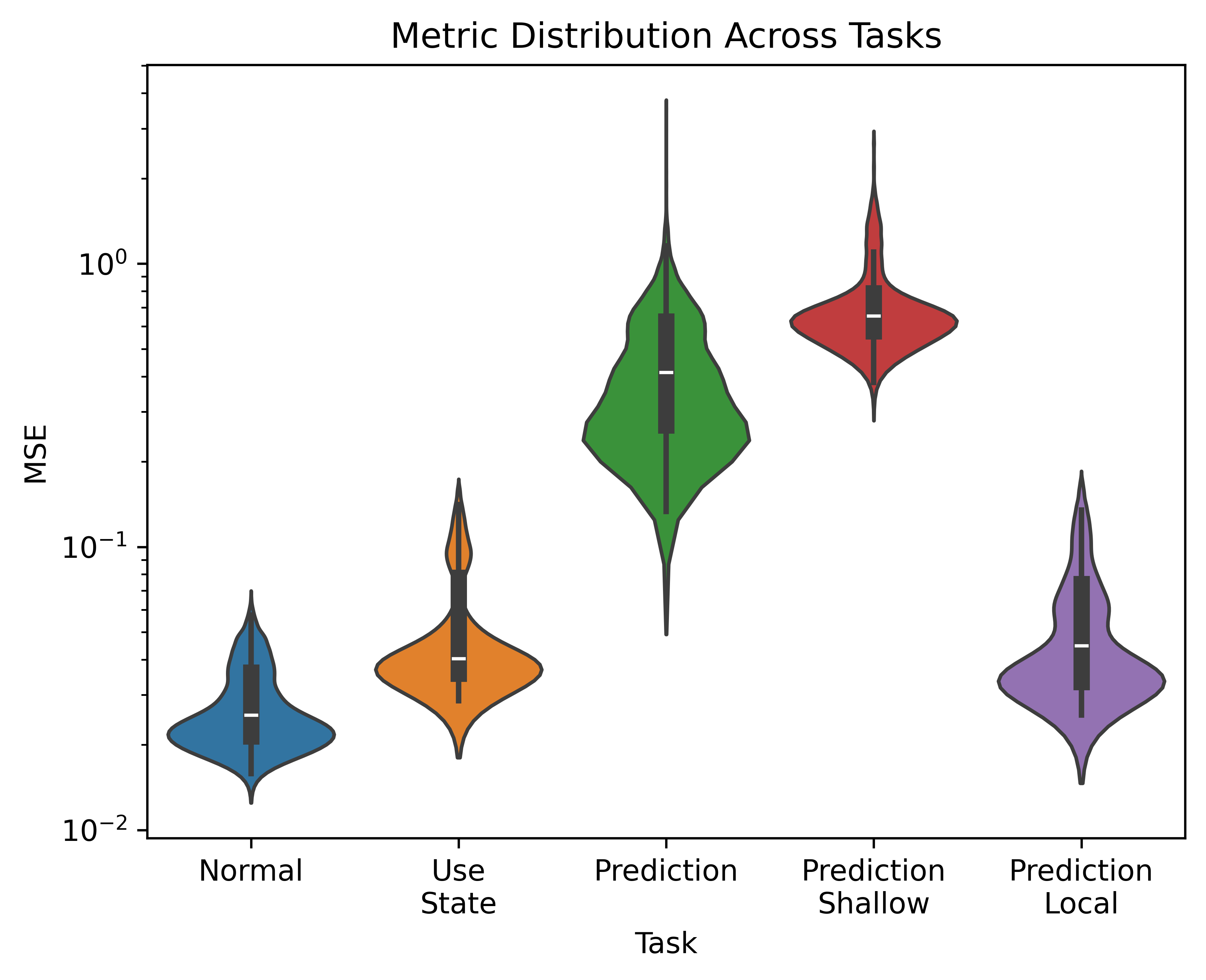}
    \caption{Metric distribution for variation of training variables}
    \label{fig:experiments-experiment1-metric_distribution}
\end{figure}

\begin{figure}
    \centering
    \includegraphics[width=0.75\linewidth]{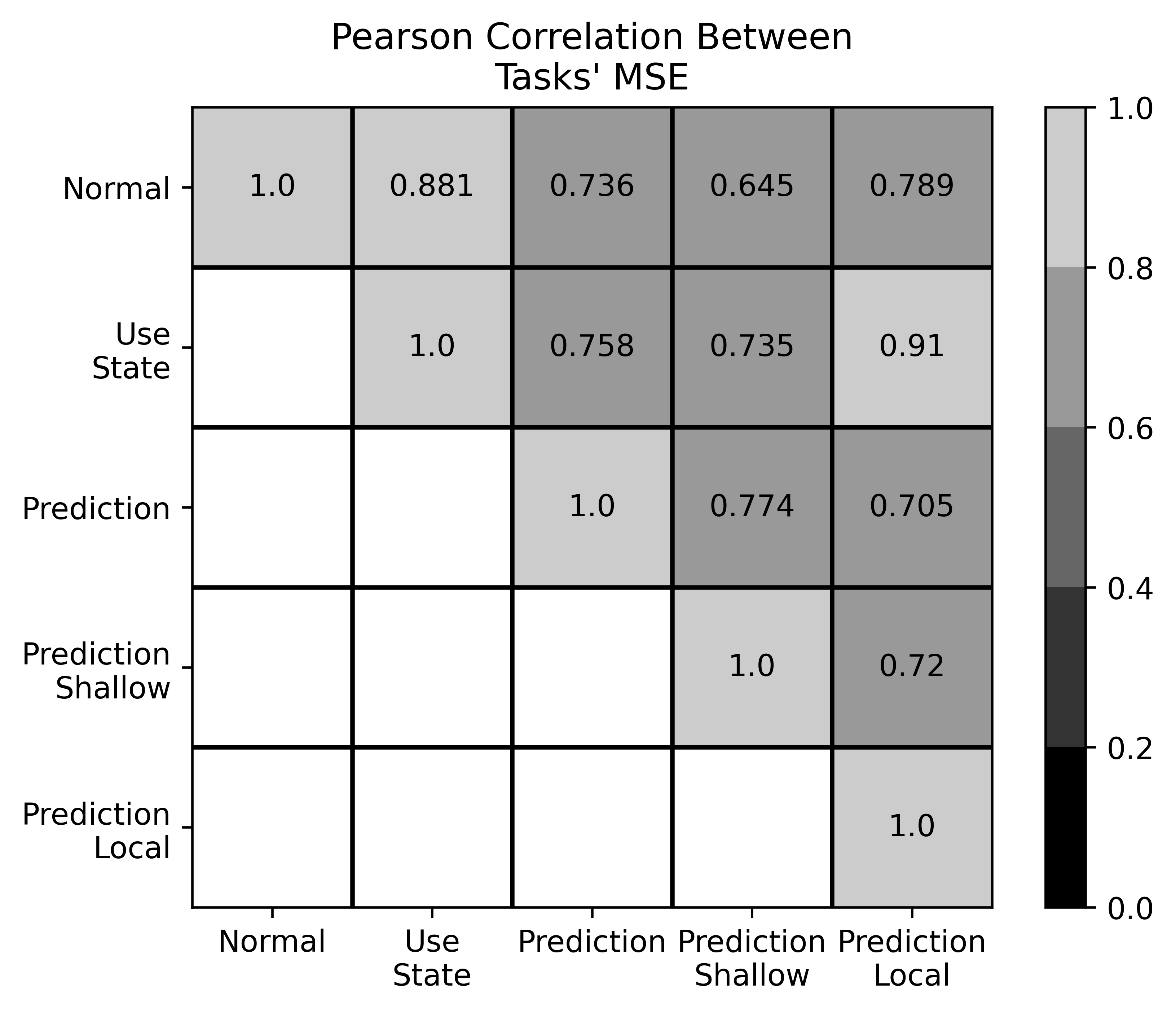}
    \caption{Correlation between task performance across variations}
    \label{fig:experiments-experiment1-correlation}
\end{figure}

Focusing on the distribution of metrics in each task (\cref{fig:experiments-experiment1-metric_distribution}), we again observed that the Prediction Shallow task is the most difficult, with the worst performance. We also observed that the distributions are skewed, with a threshold point generating a few models achieving the best observed performance. Analyzing the correlation between each task (\cref{fig:experiments-experiment1-correlation}), we observed a very high correlation between Use State and two other tasks. On the other hand, Prediction Shallow does not have very high correlation with any task. This may indicate that we can use only Use State and Prediction Shallow as evaluators of the model's performance.

\begin{figure}
    \centering
    \includegraphics[width=\linewidth]{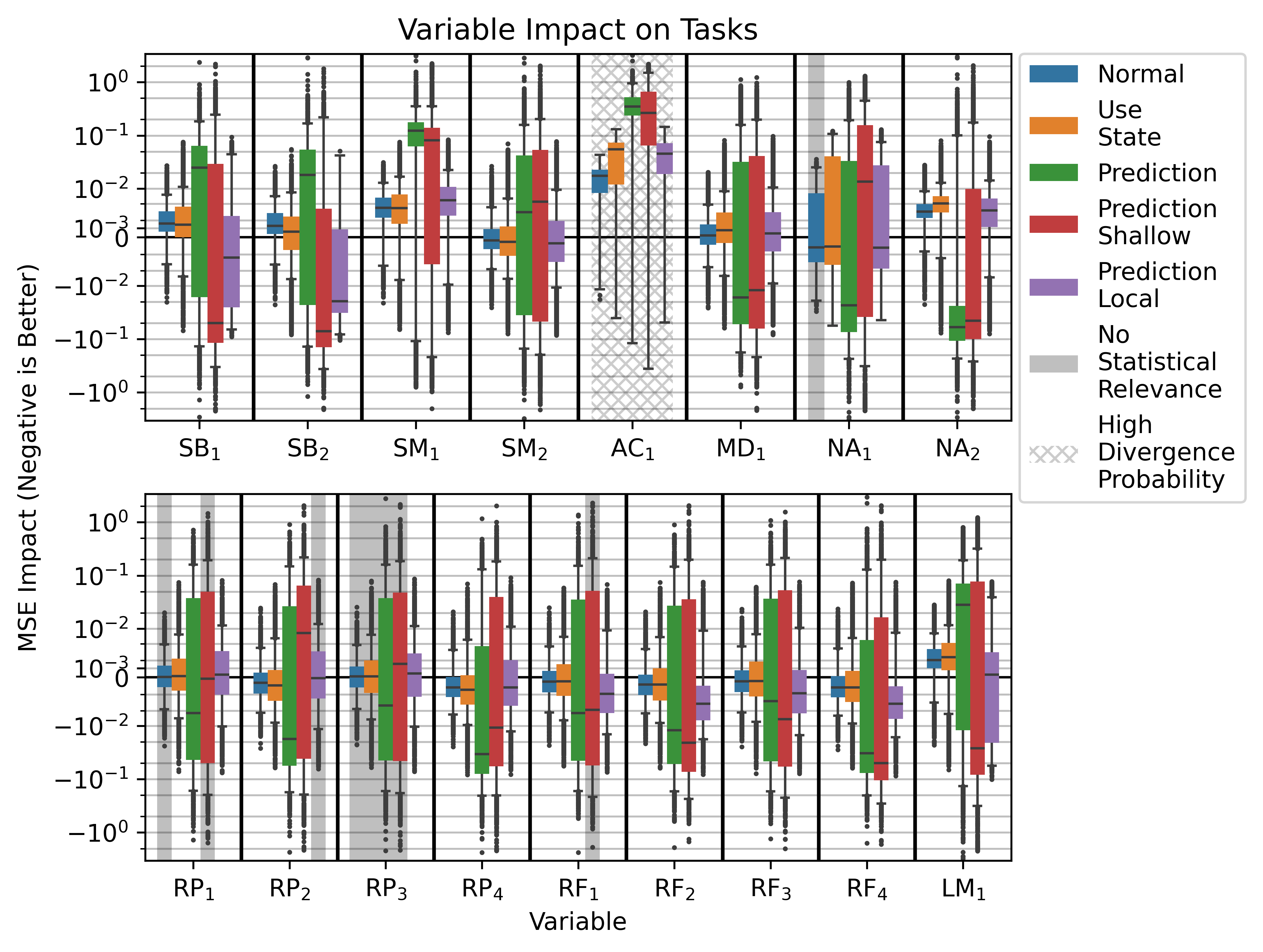}
    \caption{Impact of each training variable on task performance}
    \label{fig:experiments-experiment1-task_impact}
\end{figure}

We performed impact tests on each variable to assess its effect on performance and duration. For pairs, considering a combination of parameters with and without a specific variable, if one of the elements has diverged, we disregard the entire pair in the test. Examining the performance impact in each task (\cref{fig:experiments-experiment1-task_impact}), we first observed that the same variable can have different impacts on tasks, even in opposite directions or with different statistical significance. However, the impact tends to have high variability, making it not exclusively positive or negative. The impact on Normal and Use State tasks tends to be very small. The variable $AC_1$, which showed a high likelihood of causing divergence, is also the only one that exclusively exhibits a significant negative impact on performance in all tasks.

As the Prediction Shallow is the task with worse performance, e checked the variable configuration that would result in better performance, based on the median impact of each one and considering disjoint variables. This resulted in the configuration $\{SB_2, MD_1, NA_2, RP_4, RF_4, LM_1\}$, that empirically exhibits performance 0.4161 in the task. However, the best empirical configuration was $\{SB_2, MD_1, NA_2, RP_4, RF_4\}$, with a performance of 0.3804. Considering that the two configurations differ in only one variable, and have similar performances (which may vary due to noise), we believe that the impact obtained for each variable is a good indicator of which variables are or are not important. In this way, we can not only perform an ablation test and check the importance of each step of the training protocol, but also actually evaluate the impact of these steps and their parameters. The best empirical configuration was retrained, this time for 1,000 epochs, achieving a Prediction Shallow MSE of 0.3435, resulting in the samples in \cref{fig:main_figure} (c).

Finally, we also observed the impact of the variables on the total training time. Almost all variables showed a negative impact, with only $MD_1$ and $LM_1$ showing a positive and statistically significant impact. 
The last one is expected, since using local mode in some batches means not using the attention process, the most expensive operation in a transformer. This variable has a good impact on metrics and time, indicating a positive cost/benefit ratio. Another interesting phenomenon is the increase in the impact of $RP_{2,4}$ and $RF_{2,4}$, relative to $RP_{1,3}$ and $RF_{1,3}$. The first group creates 5 extra dimensions, while the second creates only 1. This shows how the cost is proportional to the number of dimensions created.

\begin{figure}
    \centering
    \includegraphics[width=1\linewidth]{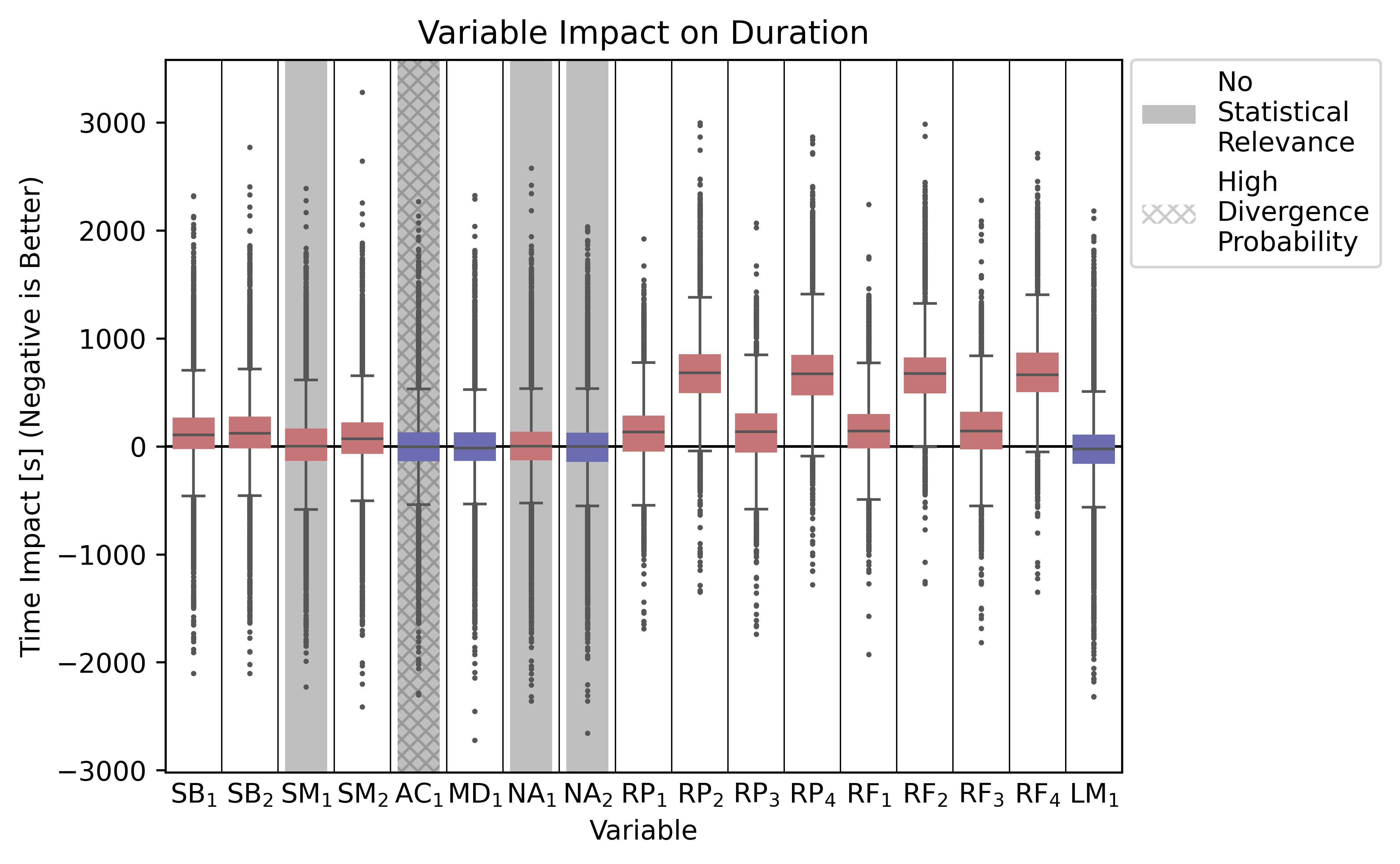}
    \caption{Training variables impact on the sum of train epoch duration. Red corresponds to a bad impact on epoch duration.}
    \label{fig:experiments-experiment1-duration_impact}
\end{figure}

\section{Discussion}
\label{sec:discussion}

Our approach, World Machine, exhibits important capabilities for creating world models, including the ability to represent the world and predict its future states. It does this adaptively, being able to handle different sensory data, which may or may not be available at each temporal step, and with a controllable amount of information about the past states. This last capability also has the potential to create more cost-effective models, given the model's ability to predict sequences from just a single element of past context, as seen in the Prediction Shallow task, although this still needs improvement. A counterpoint to this is the need for training that goes through each element of the dataset at least a few times, due to the State Discovery process, which may be impractical for massive datasets such as text datasets.

In addition to the need for improvements in the tasks presented to approach inferences without missing data, our exploration is limited by not testing the model with truly multimodal data. Future work could aim to test it on multimodal problems, such as visual-inertial odometry, or on massively multimodal datasets like Ego-Exo4D. The proposed technique can also be important in the LLM context, reducing the cost of inference and providing greater control over the cost/quality tradeoff. With tests in standard datasets, we will be able to compare the model performance with other SOA architectures. The comprehensive impact test over hyperparameters lays the foundation for training other models using this architecture, providing insights into which techniques enhance or decrease performance, along with the associated time cost of each one. 

We also draw attention to the potential ethical impacts of the work. Although it is still too early to determine what can be achieved with such an approach, current data generation models already exhibit problems, including high costs in their creation and operation, whether environmental or related to labor and data exploitation, as well as the dissemination of fake data. Continuously monitoring, reflecting on, and acting upon the potential ethical impacts is a must for the future of this work, as it cannot be a post-hoc analysis.

\vspace{3mm}
\noindent\textbf{Acknowledgments.} This project was supported by the Brazilian Ministry of Science, Technology, and Innovations, with resources from Law nº 8,248, of October 23, 1991, within the scope of PPI-SOFTEX, coordinated by Softex and published Arquitetura Cognitiva (Phase 3), DOU 01245.003479/2024 -10

{
    \small
    \bibliographystyle{ieeenat_fullname}
    \bibliography{Mestrado}

@inproceedings{cuturiSoftDTWDifferentiableLoss2017,
  title = {Soft-{{DTW}}: A Differentiable Loss Function for Time-Series},
  shorttitle = {Soft-{{DTW}}},
  booktitle = {Proceedings of the 34th {{International Conference}} on {{Machine Learning}} - {{Volume}} 70},
  author = {Cuturi, Marco and Blondel, Mathieu},
  year = 2017,
  month = aug,
  series = {{{ICML}}'17},
  pages = {894--903},
  publisher = {JMLR.org},
  address = {Sydney, NSW, Australia},
  urldate = {2025-10-31}
}

@book{molnar2025,
  title = {Interpretable Machine Learning. {{A}} Guide for Making Black Box Models Explainable},
  author = {Molnar, Christoph},
  year = 2025,
  edition = {3},
  url = {https://christophm.github.io/interpretable-ml-book},
  isbn = {978-3-911578-03-5}
}

@article{dufterPositionInformationTransformers2022,
  title = {Position {{Information}} in {{Transformers}}: {{An Overview}}},
  shorttitle = {Position {{Information}} in {{Transformers}}},
  author = {Dufter, Philipp and Schmitt, Martin and Sch{\"u}tze, Hinrich},
  year = 2022,
  month = sep,
  journal = {Computational Linguistics},
  volume = {48},
  number = {3},
  pages = {733--763},
  issn = {0891-2017, 1530-9312},
  doi = {10.1162/coli_a_00445},
  urldate = {2025-03-28},
  langid = {english}
}

@inproceedings{press2022train,
  title = {Train Short, Test Long: {{Attention}} with Linear Biases Enables Input Length Extrapolation},
  shorttitle = {Train Short, Test Long},
  booktitle = {International Conference on Learning Representations},
  author = {Press, Ofir and Smith, Noah and Lewis, Mike},
  year = 2022
}

@inproceedings{peeblesScalableDiffusionModels2023,
  title = {Scalable {{Diffusion Models}} with {{Transformers}}},
  booktitle = {2023 {{IEEE}}/{{CVF International Conference}} on {{Computer Vision}} ({{ICCV}})},
  author = {Peebles, William and Xie, Saining},
  year = 2023,
  month = oct,
  pages = {4172--4182},
  issn = {2380-7504},
  doi = {10.1109/ICCV51070.2023.00387},
  urldate = {2025-11-12},
}

@misc{worldlabsteamRTFMRealTimeFrame,
  title = {{{RTFM}}: {{A Real-Time Frame Model}}},
  shorttitle = {{{RTFM}}},
  author = {{World Labs team}},
  year = 2025,
  month = oct,
  journal = {World Labs Blog},
  url = {https://www.worldlabs.ai/blog/rtfm},
  urldate = {2025-11-03},
  langid = {english}
}

@article{kanervistoWorldHumanAction2025,
  title = {World and {{Human Action Models}} towards Gameplay Ideation},
  author = {Kanervisto, Anssi and Bignell, Dave and Wen, Linda Yilin and Grayson, Martin and Georgescu, Raluca and Valcarcel Macua, Sergio and Tan, Shan Zheng and Rashid, Tabish and Pearce, Tim and Cao, Yuhan and Lemkhenter, Abdelhak and Jiang, Chentian and Costello, Gavin and Gupta, Gunshi and Tot, Marko and Ishida, Shu and Gupta, Tarun and Arora, Udit and White, Ryen W. and Devlin, Sam and Morrison, Cecily and Hofmann, Katja},
  year = 2025,
  month = feb,
  journal = {Nature},
  volume = {638},
  number = {8051},
  pages = {656--663},
  publisher = {Nature Publishing Group},
  issn = {1476-4687},
  doi = {10.1038/s41586-025-08600-3},
  urldate = {2025-05-29},
  copyright = {2025 The Author(s)},
  langid = {english}
}

@article{decartOasisUniverseTransformer2024,
  title = {Oasis: {{A Universe}} in a {{Transformer}}},
  shorttitle = {Oasis},
  author = {Decart and Quevedo, Julian and McIntyre, Quinn and Campbell, Spruce and Wachen, Robert},
  year = 2024,
  month = oct,
  url = {https://oasis-model.github.io/},
  urldate = {2024-11-22},
  langid = {english}
}

@inproceedings{Hafner2020Dream,
  title = {Dream to Control: {{Learning}} Behaviors by Latent Imagination},
  booktitle = {International Conference on Learning Representations},
  author = {Hafner, Danijar and Lillicrap, Timothy and Ba, Jimmy and Norouzi, Mohammad},
  year = 2020,
}

@article{genie3,
  title = {Genie 3: A New Frontier for World Models},
  author = {Ball, Philip J. and Bauer, Jakob and Belletti, Frank and Brownfield, Bethanie and Ephrat, Ariel and Fruchter, Shlomi and Gupta, Agrim and Holsheimer, Kristian and Holynski, Aleksander and Hron, Jiri and Kaplanis, Christos and Limont, Marjorie and McGill, Matt and Oliveira, Yanko and {Parker-Holder}, Jack and Perbet, Frank and Scully, Guy and Shar, Jeremy and Spencer, Stephen and Tov, Omer and Villegas, Ruben and Wang, Emma and Yung, Jessica and Baetu, Cip and Berbel, Jordi and Bridson, David and Bruce, Jake and Buttimore, Gavin and Chakera, Sarah and Chandra, Bilva and Collins, Paul and Cullum, Alex and Damoc, Bogdan and Dasagi, Vibha and Gazeau, Maxime and Gbadamosi, Charles and Han, Woohyun and Hirst, Ed and Kachra, Ashyana and Kerley, Lucie and Kjems, Kristian and Knoepfel, Eva and Koriakin, Vika and Lo, Jessica and Lu, Cong and Mehring, Zeb and Moufarek, Alex and Nandwani, Henna and Oliveira, Valeria and Pardo, Fabio and Park, Jane and Pierson, Andrew and Poole, Ben and Ran, Helen and Salimans, Tim and Sanchez, Manuel and Saprykin, Igor and Shen, Amy and Sidhwani, Sailesh and Smith, Duncan and Stanton, Joe and Tomlinson, Hamish and Vijaykumar, Dimple and Wang, Luyu and Wingfield, Piers and Wong, Nat and Xu, Keyang and Yew, Christopher and Young, Nick and Zubov, Vadim and Eck, Douglas and Erhan, Dumitru and Kavukcuoglu, Koray and Hassabis, Demis and Gharamani, Zoubin and Hadsell, Raia and {van den Oord}, A{\"a}ron and Mosseri, Inbar and Bolton, Adrian and Singh, Satinder and Rockt{\"a}schel, Tim},
  year = 2025,
  url = {https://deepmind.google/blog/genie-3-a-new-frontier-for-world-models/}
}

@misc{lecunPathAutonomousMachine,
  title = {A {{Path Towards Autonomous Machine Intelligence Version}} 0.9.2, 2022-06-27},
  shorttitle = {A {{Path Towards Autonomous Machine Intelligence}}},
  author = {LeCun, Yann},
  year = 2022,
  month = jun,
  langid = {english}
}

@inproceedings{valevski2025diffusion,
  title = {Diffusion Models Are Real-Time Game Engines},
  booktitle = {The Thirteenth International Conference on Learning Representations},
  author = {Valevski, Dani and Leviathan, Yaniv and Arar, Moab and Fruchter, Shlomi},
  year = 2025
}

@article{cardosodonascimentoWorldMachineData2025,
  title = {World {{Machine}} - {{Data}} \& {{Report}} - {{Toy1D}} - {{Experiment}} 0 {{Protocol Test}}},
  author = {{Cardoso do Nascimento}, Elton and Costa, Paula},
  year = 2025,
  month = dec,
  publisher = {Zenodo},
  doi = {10.5281/zenodo.17352547},
  url = {https://zenodo.org/records/17352547},
  urldate = {2025-12-16},
  langid = {english}
}

@article{cardosodonascimentoWorldMachineData2025a,
  title = {World {{Machine}} - {{Data}} \& {{Report}} - {{Toy1D}} - {{Experiment}} 1 {{Configuration Test}}},
  author = {{Cardoso do Nascimento}, Elton and Costa, Paula},
  year = 2025,
  month = dec,
  publisher = {Zenodo},
  doi = {10.5281/zenodo.17661653},
  url = {https://zenodo.org/records/17661653},
  urldate = {2025-12-16},
  langid = {english}
}

@article{cardosodonascimentoWorldMachineData2025b,
  title = {World {{Machine}} - {{Data}} \& {{Report}} - {{Toy1D}} - {{Experiment}} 2 {{Best Long}}},
  author = {{Cardoso do Nascimento}, Elton and Costa, Paula},
  year = 2025,
  month = dec,
  publisher = {Zenodo},
  doi = {10.5281/zenodo.17653179},
  url = {https://zenodo.org/records/17653179},
  urldate = {2025-12-16},
  langid = {english}
}

@article{cardosodonascimentoWorldMachineDocker2025,
  title = {World {{Machine Docker Image}}},
  author = {{Cardoso do Nascimento}, Elton and Costa, Paula},
  year = 2025,
  month = dec,
  publisher = {Zenodo},
  doi = {10.5281/zenodo.17545425},
  url = {https://zenodo.org/records/17545425},
  urldate = {2025-12-16}
}

@misc{cardosodonascimentoWorldMachineToy1D2025,
  title = {World {{Machine}} - {{Toy1D Dataset}}},
  author = {{Cardoso do Nascimento}, Elton and Costa, Paula},
  year = 2025,
  month = dec,
  publisher = {Zenodo},
  doi = {10.5281/zenodo.17653221},
  url = {https://zenodo.org/records/17653221},
  urldate = {2025-12-16},
  langid = {english}
}

@misc{cardosodonascimentoWorldMachine2025,
  title = {World {{Machine}}},
  author = {{Cardoso do Nascimento}, Elton and Dornhofer Paro Costa, Paula},
  year = 2025,
  month = dec,
  doi = {10.5281/zenodo.17806741},
  url = {https://github.com/H-IAAC/WorldMachine},
  urldate = {2025-12-16},
  howpublished = {Zenodo}
}
}

\clearpage
\setcounter{page}{1}
\maketitlesupplementary
\label{sec:supplementary}

\begin{figure}[tp]
  \centering
    \begin{subfigure}[t]{\linewidth}
        \includegraphics[width=\linewidth]{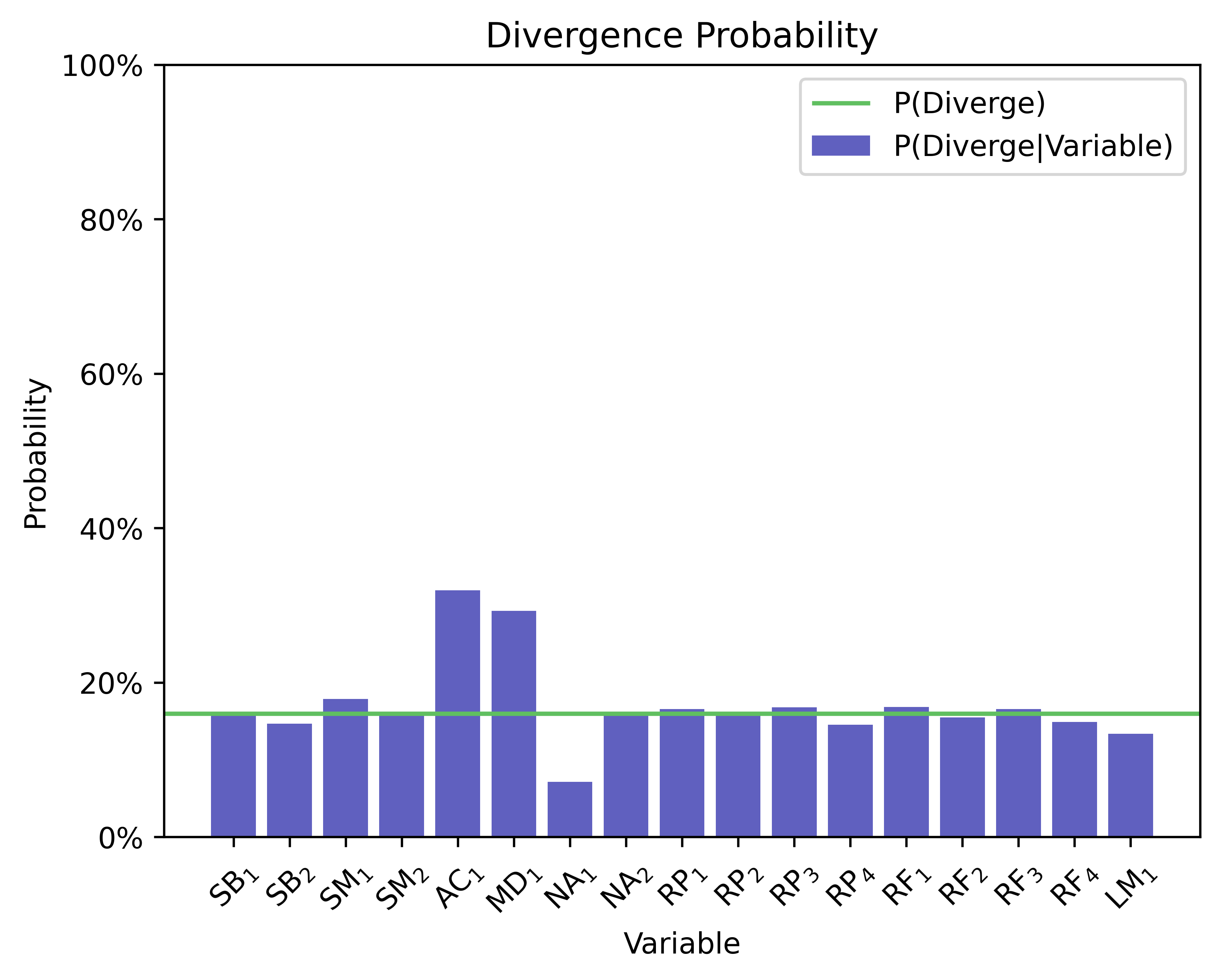}
        \caption[b]{All variables}
        \label{fig:experiments-experiment1-divergence-unfiltered}
    \end{subfigure}
    
    \begin{subfigure}[t]{\linewidth}
        \includegraphics[width=\linewidth]{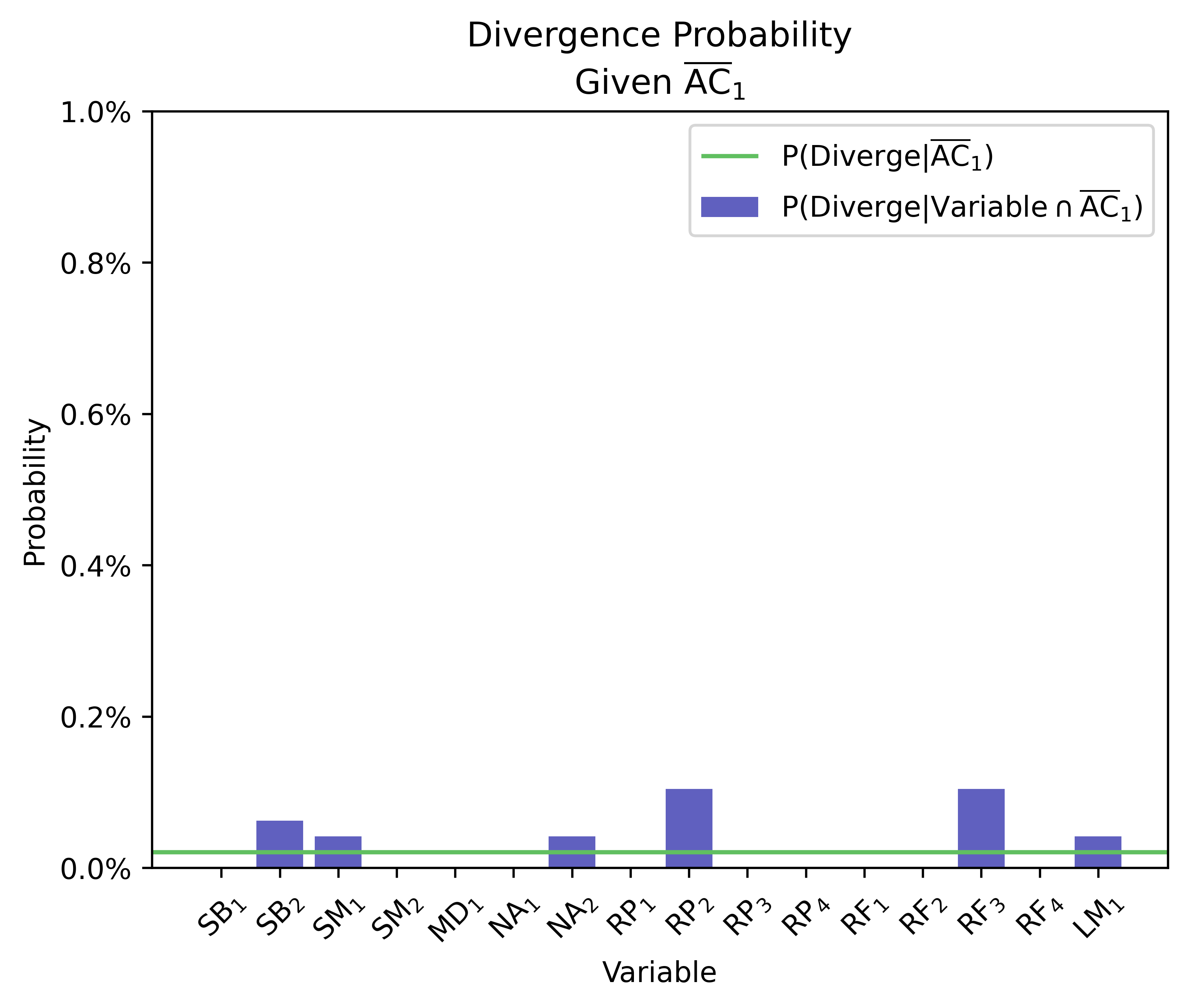}
        \caption[b]{Excluding variable \protect{$AC_1$}}
        \label{fig:experiments-experiment1-divergence-filtered}
    \end{subfigure}
    \caption{Divergence probability conditioned by training variable}
    \label{fig:experiments-experiment1-divergence}
\end{figure}

\Cref{fig:experiments-experiment1-divergence-unfiltered} shows the divergence probability conditioned by each variable. We can observe a considerable chance of divergence. However, when disregarding variations that used the variable $AC_1$ (\cref{fig:experiments-experiment1-divergence-filtered}), we observe a much smaller chance of divergence.

---

\cref{tab:experiments-experiment1-variables} shows all the defined variables.

\begin{table*}[]
    \caption{Training variable relation with hyperparameters. For the \protect\say{Sequence Breaking}, \protect\say{Short Time Recall - Future} and \protect\say{Short Time Recall - Past} domains, only one variable can be active. The \protect\say{BASE} variable refers for the model trained without any variable in a domain. \protect\say{M} and \protect\say{S} refers to the sensorial dimension \protect\say{Measurement} and \protect\say{State on the core input}, respectively.}
    \label{tab:experiments-experiment1-variables}
    \centering
\begin{tabular}{lllll}
\hline
\textbf{Variable Domain}                     & \textbf{Variable} & \multicolumn{3}{l}{\textbf{Hyperparameter combination}}                                                           \\ \hline
                                             &                   & \textbf{n\_segment}           & \textbf{fast\_forward}          & \cellcolor[HTML]{EFEFEF}{\color[HTML]{C0C0C0} } \\
                                             & BASE              & \multicolumn{1}{r}{1}         & \multicolumn{1}{c}{FALSE}       & \cellcolor[HTML]{EFEFEF}{\color[HTML]{C0C0C0} } \\
                                             & SB\_1             & \multicolumn{1}{r}{2}         & \multicolumn{1}{c}{FALSE}       & \cellcolor[HTML]{EFEFEF}{\color[HTML]{C0C0C0} } \\
\multirow{-4}{*}{Sequence Breaking}          & SB\_2             & \multicolumn{1}{r}{2}         & \multicolumn{1}{c}{TRUE}        & \cellcolor[HTML]{EFEFEF}{\color[HTML]{C0C0C0} } \\ \hline
                                             &                   & \textbf{state\_save\_method}  & \textbf{check\_input\_masks}    & \cellcolor[HTML]{EFEFEF}{\color[HTML]{C0C0C0} } \\
                                             & BASE              & REPLACE                       & \multicolumn{1}{c}{FALSE}       & \cellcolor[HTML]{EFEFEF}{\color[HTML]{C0C0C0} } \\
                                             & SM\_1             & MEAN                          & X                               & \cellcolor[HTML]{EFEFEF}{\color[HTML]{C0C0C0} } \\
\multirow{-4}{*}{State Discovery}            & SM\_2             & X                             & \multicolumn{1}{c}{TRUE}        & \cellcolor[HTML]{EFEFEF}{\color[HTML]{C0C0C0} } \\ \hline
                                             &                   & \textbf{state\_activation}    & \textbf{state\_regularizer}     & \cellcolor[HTML]{EFEFEF}{\color[HTML]{C0C0C0} } \\
                                             & BASE              & tanh                          & NONE                            & \cellcolor[HTML]{EFEFEF}{\color[HTML]{C0C0C0} } \\
\multirow{-3}{*}{State Activation}           & AC\_1             & NONE                          & MSE                             & \cellcolor[HTML]{EFEFEF}{\color[HTML]{C0C0C0} } \\ \hline
                                             &                   & \textbf{block\_configuration} & \cellcolor[HTML]{EFEFEF}        & \cellcolor[HTML]{EFEFEF}                        \\
                                             & BASE              & {[}M, M{]}                    & \cellcolor[HTML]{EFEFEF}        & \cellcolor[HTML]{EFEFEF}                        \\
\multirow{-3}{*}{Model Block Configuration}  & MD\_1             & {[}M, S{]}                    & \cellcolor[HTML]{EFEFEF}        & \cellcolor[HTML]{EFEFEF}                        \\ \hline
                                             &                   & \textbf{noise\_config}        & \cellcolor[HTML]{EFEFEF}        & \cellcolor[HTML]{EFEFEF}                        \\
                                             & BASE              & None                          & \cellcolor[HTML]{EFEFEF}        & \cellcolor[HTML]{EFEFEF}                        \\
                                             & NA\_1             & state, X                      & \cellcolor[HTML]{EFEFEF}        & \cellcolor[HTML]{EFEFEF}                        \\
\multirow{-4}{*}{Noise Addition}             & NA\_2             & X, measurement                & \cellcolor[HTML]{EFEFEF}        & \cellcolor[HTML]{EFEFEF}                        \\ \hline
                                             &                   & \textbf{short\_time\_recall}  & \textbf{recall\_stride\_future} & \textbf{recall\_n\_future}                      \\
                                             & BASE              & None                          & X                               & X                                               \\
                                             & RF\_1             & M                             & \multicolumn{1}{r}{1}           & \multicolumn{1}{r}{1}                           \\
                                             & RF\_2             & M                             & \multicolumn{1}{r}{1}           & \multicolumn{1}{r}{5}                           \\
                                             & RF\_3             & M                             & \multicolumn{1}{r}{3}           & \multicolumn{1}{r}{1}                           \\
\multirow{-6}{*}{Short Time Recall - Future} & RF\_4             & M                             & \multicolumn{1}{r}{3}           & \multicolumn{1}{r}{5}                           \\ \hline
                                             &                   & \textbf{short\_time\_recall}  & \textbf{recall\_stride\_past}   & \textbf{recall\_n\_past}                        \\
                                             & BASE              & None                          & X                               & X                                               \\
                                             & RP\_1             & M                             & \multicolumn{1}{r}{1}           & \multicolumn{1}{r}{1}                           \\
                                             & RP\_2             & M                             & \multicolumn{1}{r}{1}           & \multicolumn{1}{r}{5}                           \\
                                             & RP\_3             & M                             & \multicolumn{1}{r}{3}           & \multicolumn{1}{r}{1}                           \\
\multirow{-6}{*}{Short Time Recall - Past}   & RP\_4             & M                             & \multicolumn{1}{r}{3}           & \multicolumn{1}{r}{5}                           \\ \hline
                                             &                   & \textbf{local\_chance}        & \cellcolor[HTML]{EFEFEF}        & \cellcolor[HTML]{EFEFEF}                        \\
                                             & BASE              & None                          & \cellcolor[HTML]{EFEFEF}        & \cellcolor[HTML]{EFEFEF}                        \\
\multirow{-3}{*}{Local Mode}                 & LM\_1             & 0.25                          & \cellcolor[HTML]{EFEFEF}        & \cellcolor[HTML]{EFEFEF}                        \\ \hline
\end{tabular}
\end{table*}

\end{document}